\documentclass[sigconf]{acmart}

\usepackage{amsmath,amsfonts}
\usepackage{graphicx}
\usepackage{textcomp}
\usepackage{subfig}

\copyrightyear{2020}
\acmYear{2020}
\setcopyright{acmlicensed}\acmConference[FDG '20]{International Conference on the Foundations of Digital Games}{September 15--18, 2020}{Bugibba, Malta}
\acmBooktitle{International Conference on the Foundations of Digital Games (FDG '20), September 15--18, 2020, Bugibba, Malta}
\acmPrice{15.00}
\acmDOI{10.1145/3402942.3409598}
\acmISBN{978-1-4503-8807-8/20/09}

\begin{document}

\title{10 Years of the PCG workshop: Past and Future Trends}
\author{Antonios Liapis}
\affiliation{%
  \institution{Institute of Digital Games, University of Malta}
  \city{Msida}
  \country{Malta}
}
\email{antonios.liapis@um.edu.mt}

\begin{abstract}
As of 2020, the international workshop on Procedural Content Generation enters its second decade. The annual workshop, hosted by the international conference on the Foundations of Digital Games, has collected a corpus of 95 papers published in its first 10 years. This paper provides an overview of the workshop's activities and surveys the prevalent research topics emerging over the years.
\end{abstract}



\maketitle

\section{Introduction}\label{sec:introduction}

While algorithms have been applied in digital games for almost 40 years, in games such as \emph{Rogue} (Toy and Wichman, 1981) and \emph{ELITE} (Acornsoft, 1984), their prominence as a field of academic research is fairly recent \cite{togelius2011searchbased}. The workshop on Procedural Content Generation (PCG) has served an important role in formalizing the field of study, collecting publications, and mobilizing an academic community interested in PCG research. Since its first incarnation in 2010, the PCG workshop has been hosted by the international conference on the Foundations of Digital Games (FDG). At the time of writing, May 2020, the workshop has been running consecutively for ten years. As a celebration of the 11th anniversary of the PCG workshop in 2020, this paper looks back at the workshops' ten past iterations and identifies topics and directions in its 95 published papers. The survey covers all published papers at the ten first PCG workshops, collected from the ACM Digital Library and the workshops' websites in years where the workshop proceedings were not published by ACM.

In the decade since the first PCG workshop, research in artificial intelligence (AI) for generating game content has bloomed. PCG research of all types has been accepted in high-tier conferences and journals, and three special issues on topics directly relevant to PCG  \cite{togelius2011specialissue,browne2012specialissue,liapis2019specialissue} were published in the IEEE Transactions on Games (and the preceding IEEE Transactions on Computational Intelligence and AI in Games). A textbook on Procedural Content Generation in games was published \cite{shaker2016procedural}, and a Dagstuhl seminar took place in 2018 specifically focusing on AI-driven game design \cite{spronck2018dagstuhl}. Finally, the annual PROCJAM event\footnote{\url{https://itch.io/jam/procjam}} has been running since 2014 and has fostered a community of developers and researchers interested in ``making something that makes something''. Despite these developments---or perhaps because of them---the PCG workshop remains a highly visible academic venue for publishing PCG research and for researchers to pitch and discuss ideas, algorithms, and games.

\begin{table*}[t]
\centering
\begin{tabular}{|l|l|p{8cm}|l|c|}
\hline
Year & Date & Organizers & Location & \# papers \\ \hline
2010 & June 18 & Rafael Bidarra, Ian Bogost, Ian Parberry, Kenneth O. Stanley, Julian Togelius, Jim Whitehead, R. Michael Young & Monterey, USA & 11 \\ \hline
2011 & June 28 & Joris Dormans, Michael Mateas, Ian Parberry, Gillian Smith, Julian Togelius, Jim Whitehead, R. Michael Young & Bordeaux, France & 9 \\ \hline
2012 & May 29 & Julian Togelius, Joris Dormans & Raleigh, USA & 13 \\ \hline
2013 & May 15 & Alex Pantaleev, Gillian Smith, Joris Dormans, Antonio Coelho & Chania, Greece & 7 \\ \hline
2014 & April 4 & Noor Shaker, Kenneth O. Stanley, Kate Compton & Royal Caribbean Liberty of the Seas & 5 \\ \hline
2015 & June 23 & Noor Shaker, Antonios Liapis, Sebastian Risi & Monterey, USA & 11 \\ \hline
2016 & August 1 & Rafael Bidarra, Amy K. Hoover, Aaron Isaksen & Dundee, Scotland & 10 \\ \hline
2017 & August 14 & Chris Martens, Adam Summerville, Tommy Thompson & Hyannis, USA & 10 \\ \hline
2018 & August 7 & Gabriella Barros, Maria Teresa Llano, Anne Sullivan & Malm\"o, Sweden & 9 \\ \hline
2019 & August 27 & April Grow, Ahmed Khalifa and Sam Snodgrass & San Luis Obispo, USA & 10 \\ \hline
\end{tabular}
\caption{Overview of the ten years of the PCG workshop}
\label{tab:overview}
\end{table*}

\section{PCG workshops of the last decade}\label{sec:general_metrics}

Table \ref{tab:overview} lists the general information regarding the first ten PCG workshops examined in this paper. As noted in Section \ref{sec:introduction}, the first PCG workshop took place in 2010 during the FDG conference at Monterey, California. In the first four years the organizational committee was fairly stable, and a subset of the 1st and 2nd workshops' organizers helped run the next workshops. Joris Dormans, Julian Togelius and Gillian Smith were organizers in three of these first PCG workshops, each time with a different set of co-organizers. After 2014, a rotation of organizers was established to ensure that the workshop would not become stale and a larger part of the research community would be engaged in organizational tasks.

In terms of output, the PCG workshop has had a fairly stable number of papers published annually, with little fluctuation. An outlier is the 2014 workshop with only five papers published; this may be due to the fact that FDG 2014 took place on a cruise ship and thus the logistics and cost made attendance only for a workshop paper prohibiting. The last five years have been very stable in terms of participation, with a community of 30-40 people usually attending and on average 10 papers published every year.

\begin{figure*}[p]
    \centering
    \includegraphics[width=0.9\textwidth]{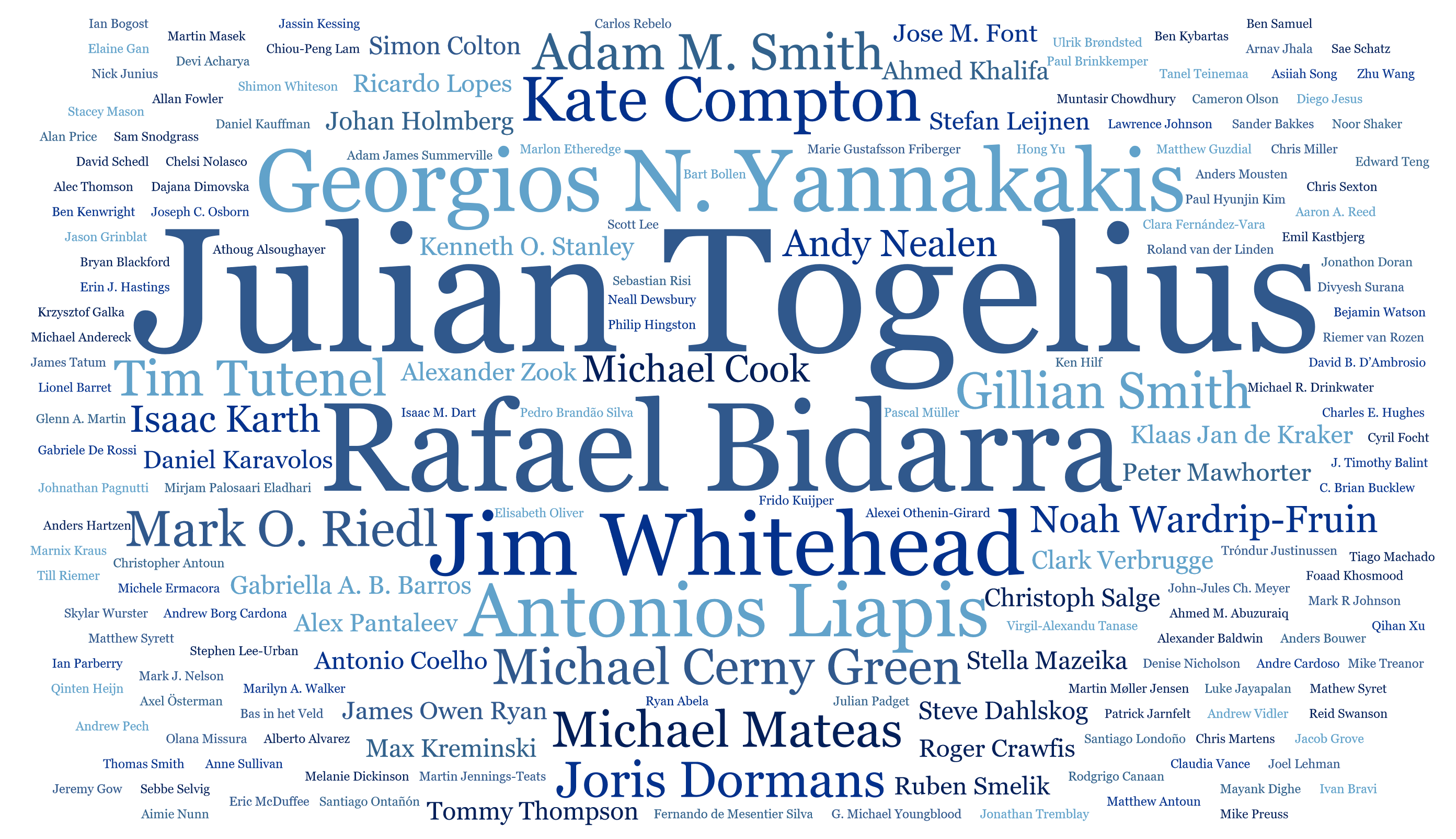}
    \caption{A word cloud of all authors in published papers at the PCG workshop. The size is proportionate to the number of co-authored papers.}
    \label{fig:wordcloud_authors}
\end{figure*}

On a surface-level analysis of the 95 published PCG papers over the course of the workshop's ten years, papers are authored on average by 2.9 co-authors (median: 3). The corpus of 95 papers includes 17 single-author papers and 10 papers with five or more authors. Over the ten years examined, 184 distinct authors have contributed to the published papers. Fig.~\ref{fig:wordcloud_authors} shows all authors of published PCG papers; it should be noted that 78\% of authors contributed only to one paper while Julian Togelius and Rafael Bidarra were the most active authors with 14 and 10 papers respectively.

\begin{figure*}[p]
    \centering
    \includegraphics[width=0.9\textwidth]{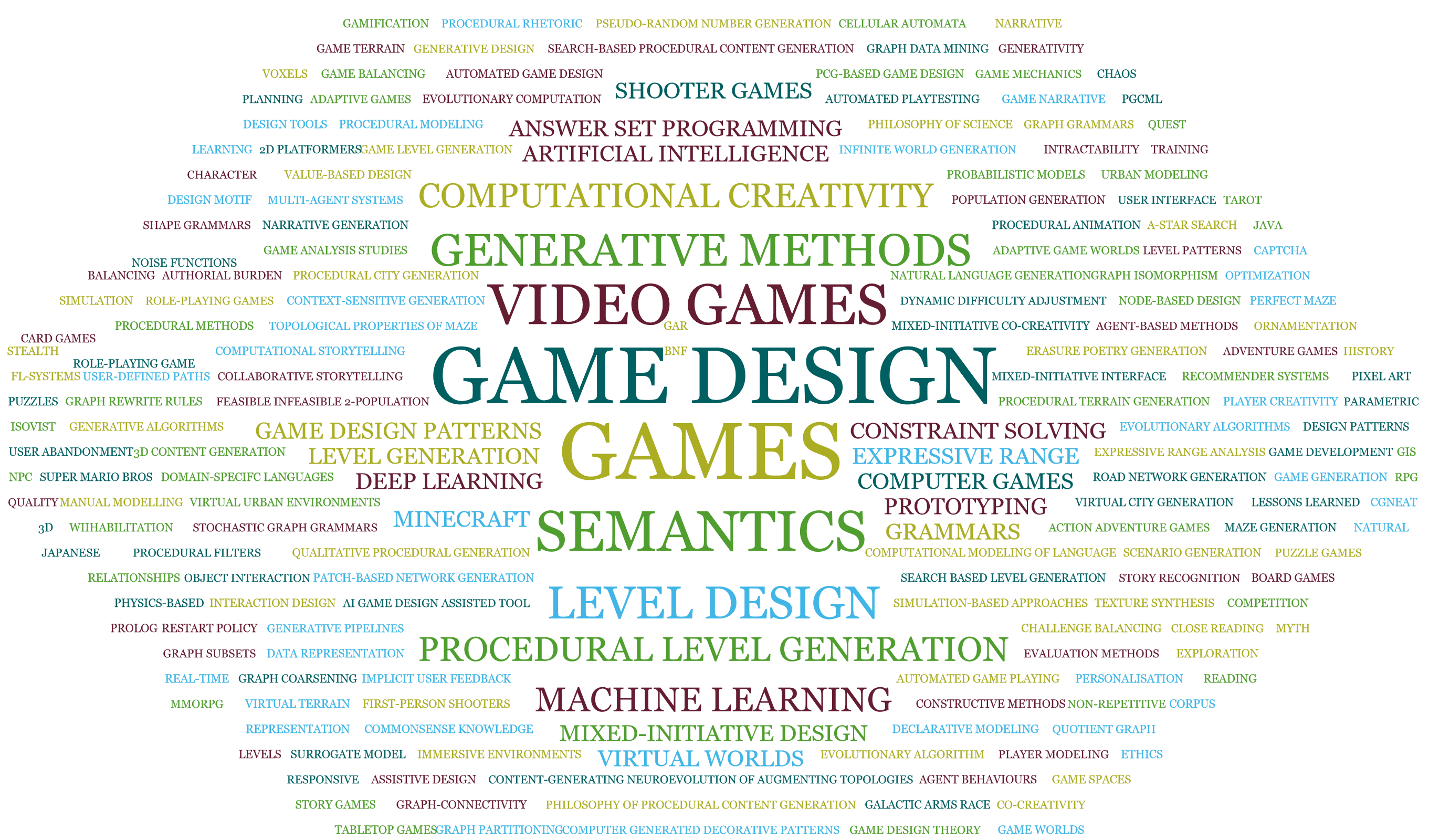}
    \caption{A word cloud of keywords in published papers at the PCG workshop. The size is proportionate to the number of appearances in published papers. Note that the generic keywords ``procedural content generation'', ``procedural generation'', ``procedural content'' and ``PCG'' have been omitted.}
    \label{fig:wordcloud_keywords}
\end{figure*}

The keywords provided by authors could also shed some light on the topics favored by the PCG workshop community. However, it should be noted that 20\% of papers did not include any keywords. Of the keywords found, Fig.~\ref{fig:wordcloud_keywords} shows their distribution while omitting the generic keywords ``procedural content generation'' (in 42 instances), ``procedural generation'' (in 7 instances), ``procedural content'' (in 2 instances), ``PCG'' (in 1 instance). Due to the lack of keywords for a fifth of the corpus, and the fairly ad-hoc way in which keywords are chosen by each author, the thematic analysis performed in the following sections contains richer information regarding the topics covered by the 95 PCG papers.

\section{Types of content generated}\label{sec:generated_content}

PCG refers to game content in terms of ``all aspects of the game that affect gameplay other than non-player character (NPC) behaviour and the game engine itself'' \cite{togelius2011searchbased} and includes ``such aspects as terrain, maps, levels, stories, dialogue, quests, characters, rulesets, dynamics and weapons'' \cite{togelius2011searchbased}. 
While game content is fairly broadly defined and incorporates most facets of games \cite{liapis2014gamecreativity}, the majority of PCG workshop papers focus on the generation of levels or worlds (51\%). The category of level generation is fairly broad, and encompasses the generation of terrain \cite{smelik2010integrating,pech2016identifying,compton2012anza}, dungeons \cite{dormans2010adventures,alvarez2018fostering,cerny2019two-step}, mazes \cite{hyunjin2019design-centric}, settlements \cite{sexton2010vectorization,in2015procedural,salge2018generative,song2019townsim:}, and many others. Tightly related to but not subsumed by level generation, architecture is also a common type of generated content: six papers published in the first decade of the PCG workshop focus on architecture, e.g. as 3D models of buildings, as interior placement of furniture \cite{timothy2019a} or decoration \cite{whitehead2010toward}. Often, the distinction between architecture and level design is difficult; for instance, settlement generation \cite{salge2018generative,in2015procedural} is attributed to the latter while texture changes on a building \cite{tutenel2011procedural} is grouped under architecture. The main criteria for the classification as generated architecture was based on fidelity with real-world patterns and the focus on aesthetic appeal rather than gameplay. 

Among other types of generated content, graphics are also prominent (in 8\% of papers), and include particle effects \cite{j.2010interactive}, 3D meshes \cite{abela2015a,m.2011speedrock:} and 2D art \cite{karth2017wavefunctioncollapse, risi2014automatically}. The generation of mechanics is equally popular (8\% of papers), sometimes combined with level generation \cite{dormans2011level} or graphics generation \cite{mcduffee2013team}. Mechanics refer to actions players can take \cite{dormans2011level,zook2014generating}, but in this case a broader view of mechanics is used to include gameplay parameters of weapons and character classes \cite{mcduffee2013team,karavolos2018pairing,pantaleev2012in}. Generated story elements also feature in 5 papers, although the classification is broad since papers range from generating plots for mysteries \cite{kreminski2019cozy}, to NPC biographies \cite{grinblat2017subverting}, to satellite sentences \cite{a.2012sharing}. Quests or missions are generated in 6 papers, either as a mission graph for a level \cite{dormans2010adventures,dormans2012generating} or for challenges in serious games \cite{a.2010the,olson2015teaching}, quests in RPGs \cite{doran2011a}, etc. The generation of full games is also tackled in 3 papers, two of which present actual game generators \cite{treanor2012game-o-matic:,a.2015data}.

Quite a few papers (12) propose generators that do not fall into the above categories: among those, some papers attempt to generate other aspects of games such as character locomotion \cite{kenwright2012generating}, language \cite{r2016procedural}, CAPTCHAs for human computation \cite{yu2015automatic} or LEGO structures \cite{mazeika2015design}. Other papers propose general algorithms without a specific type of content in mind, e.g. for graph matching \cite{etheredge2012fast}, random number generation \cite{mawhorter2019anarchy:}, or family tree representation \cite{mawhorter2017eficiency}. Finally, 5 papers do not focus on any domain but instead propose new general generation methods \cite{compton2013generative,gustafsson2013data}, or address philosophical issues such as the intentions of generators' coders \cite{smith2017what,compton2017litle} and the research field of PCG as a whole \cite{whitehead2017art}. 

\section{Target Games}\label{sec:target_games}

Looking at the types of games targeted in the research published in the PCG workshop, Table \ref{tab:game_genres} attempts a classification of the work. It should be noted that five papers covered multiple game genres while 19 papers do not target a specific genre, and this will be discussed below. 

Unsurprisingly, popular genres are platformers, roguelikes and RPGs. \emph{Super Mario Bros.} (Nintendo, 1985) is the most common platformer being targeted \cite{togelius2011what,bakkes2014towards,londono2015graph,guzdial2015toward,james2016the,cerny2018generating}, but custom platformers have also been created by researchers either as commercial games, e.g. \emph{Sure Footing} \cite{tompson2018play,dewsbury2016scalable}, or as demonstrators of player-controlled PCG \cite{cook2016towards,smith2011pcg-based}. In the context of this survey, roguelikes cover many tile-based dungeon-crawling games similar to \emph{Diablo} (Blizzard, 1997) as well as action adventure games \cite{dormans2011level,dormans2012generating,dormans2010adventures,smith2018graph-based,van2018measuring}. As expected, the generation of dungeons \cite{togelius2012compositional,cerny2019two-step,alvarez2018fostering,baldwin2017towards} and caves \cite{johnson2010cellular} is a common trope of this genre. Role playing games (RPGs) also cover a broad range of games, and have been targeted in PCG workshop papers for different purposes including quest generation \cite{doran2011a,zook2012skill-based,olson2015teaching}, class balancing \cite{pantaleev2012in}, as well as the simulation of societies \cite{in2015procedural}, their histories \cite{grinblat2017subverting} and language \cite{r2016procedural,ryan2016diegetically}.

Among other genres mentioned in Table \ref{tab:game_genres}, worth noting are serious games and real-world simulations. ``Serious'' games, or games with a purpose beyond entertainment, have been targeted by PCG workshop papers in the context of language learning \cite{olson2015teaching}, rehabilitation \cite{dimovska2010towards,andereck2013procedural}, training scenarios \cite{a.2010the,leijnen2015generating} and human computation \cite{yu2015automatic}. Real-world simulations have also been common, simulating realistic urban planning \cite{sexton2010vectorization,teng2017a,song2019townsim:,jesus2012modeling}, architecture \& interior design \cite{tutenel2011procedural,brandao2013node-based,timothy2019a}, environments \cite{abela2015a}, locomotion \cite{kenwright2012generating} or economies \cite{leijnen2015generating}.

Five papers encompass more than one genre: two papers combine serious games with another genre \cite{leijnen2015generating,antoun2015generating}, two papers apply the same algorithm on multiple genres \cite{zook2014generating,chowdhury2016exhaustive}, while \cite{james2016the} introduces a level corpus for different kinds of games (RPGs, roguelikes, arcade games and shooters). On the other hand, 19 papers also are genre-agnostic, primarily vision papers or papers on design tools, graphics, and general algorithms.

\begin{table}[t]
\centering
\begin{tabular}{|l|c|}
\hline
\textbf{Type of game} & \textbf{\# of papers} \\
\hline
Platformers	&	14	\\
Real-world simulation	&	13	\\
RPGs	&	12	\\
Roguelikes &	12	\\
Serious games	&	6	\\
Shooters	&	6	\\
Casual	&	5	\\
Arcade	&	4	\\
Tabletop	&	3	\\
Crafting	&	3	\\
Adventure	&	2	\\
Driving	&	2	\\
Strategy	&	1	\\
Stealth	&	1	\\
Other	&	1	\\
\hline
\end{tabular}
\caption{Game genres targeted in PCG workshop papers}
\label{tab:game_genres}
\end{table}

\section{Algorithmic Processes}\label{sec:algorithms}
In the context of academic trends, perhaps the most relevant aspect of the PCG workshop papers is the algorithm used for generation. However, in many cases there is no clear-cut classification for such algorithms. We revisit this later in this Section, and in Section \ref{sec:discussion}.

In the last decade, 14 papers published at the PCG workshop use artificial evolution to generate levels (8 papers) and/or mechanics (4 papers), or graphics \cite{mcduffee2013team,j.2010interactive,m.2011speedrock:}. Machine learning has also been applied in 7 PCG workshop papers, in the form of non-negative matrix factorization \cite{zook2012skill-based}, random forests \cite{bakkes2014towards}, self-organizing maps \cite{risi2014automatically}, deep learning \cite{karavolos2017learning,karavolos2018pairing}, and wave function collapse \cite{karth2017wavefunctioncollapse,karth2019addressing}. Notably, 4 of these 7 papers are published from 2017 onwards. Declarative programming is also prominent, in 13 papers, including semantic constraints 
(5 papers) and answer-set programming (8 papers). Of these papers, 62\% are published in the first 3 years of the workshop.
15 PCG workshop papers use some sort of grammar, either as templates for sentence/language generation \cite{a.2012sharing,ryan2016diegetically,sullivan2018tarot-based,miller2019stories} or for more traditional PCG goals such as level generation (4 papers), architecture \cite{brandao2013node-based}, design \cite{mazeika2015design}, or art \cite{pagnutti2016do}. However, the most common algorithmic approach in PCG papers of the last decade were constructive methods \cite{shaker2016constructive} which include industry-standard scripts such as cellular automata, agents, L-systems etc. Many papers in this corpus could qualify as using constructive methods, since this broad category likely encompasses anything that does not fall into the other better-defined families of algorithms. Even with stricter criteria, constructive methods are found in 14 papers, including cellular automata (7 papers), L-systems \cite{a.2010the,abela2015a}, recursive subdivision \cite{m.2011two}, diamond-square \cite{andereck2013procedural}, agents \cite{teinemaa2015a,song2019townsim:}. Should we have included all scripted methods, the number of PCG papers that belong to this category wound increase dramatically, but would make classification difficult since ultimately any algorithm is a scripted method.

\section{Other Aspects}\label{sec:other}

Looking at other trends in the PCG workshop papers of the last decade, it is evident that design tools (for or with PCG) are popular (in 19 papers). Some papers focus on presenting the tool itself \cite{smelik2010integrating,barret2011lessons,machado2016shopping,cook2016danesh:,baldwin2017towards,alvarez2018fostering} while most papers of this type focus on the algorithms and briefly expose a way for the designer to control these algorithms.

Other popular topics include the use of design patterns to help guide generation (7 papers), usually for level generation \cite{dewsbury2016scalable,baldwin2017towards,alvarez2018fostering}, the use of expressive range as an evaluation method (7 papers), and the real-time adaptation/modification of game content while the game is played (7 papers). It should be noted that expressive range as a concept was introduced in the first PCG workshop by Smith and Whitehead \cite{smith2010analyzing} as a method for evaluating level generators: indeed, all 7 PCG papers that use expressive range analysis apply it to level generation.

Of interest is also the fact that modelling the players' behavior or progress is attempted in 5 PCG workshop papers, all of which are published in the first 5 years of the workshop. Modelling is performed primarily to adapt the game's difficulty or challenge \cite{bakkes2014towards,jennings-teats2010polymorph:,zook2012skill-based,lopes2013mobile} and to match players' preferences \cite{lopes2012using}. Close to the latter goal of modelling, 5 PCG workshop papers place content generation at the hands of the player, as part of the game mechanics \cite{cook2016towards} and/or a direct response to player actions \cite{togelius2011what,compton2012anza,smith2011pcg-based}. Cases where players control the generator's settings \cite{tompson2018play} are more difficult to classify as their distinction from designer tools is not clear-cut, e.g. in Game-O-Matic \cite{treanor2012game-o-matic:}.

Finally, it is important to note that 8 PCG papers introduce a vision and offer little technical details, proposing new types of content for PCG \cite{whitehead2010toward}, new general methods for generation \cite{gustafsson2013data,compton2013generative,kreminski2019generators} or expose ethical or general issues that PCG is facing \cite{togelius2014characteristics,smith2017what,whitehead2017art,compton2017litle}. Beyond these vision papers, 92\% of papers published at the PCG workshop contain technical contributions, e.g. introducing new algorithms, corpora, games, methods, and a competition \cite{cerny2018generating}. Despite the focus on technical contributions, only 32 papers include an empirical evaluation. We define ``evaluation'' as a quantitative analysis of multiple runs of the generator (with or without human testers) which reports aggregated statistics and/or performs significance tests. Instead, most surveyed papers include a couple of generated samples and a qualitative discussion on the workings of the system. Quantitative evaluation is primarily applied on level generation tasks (87\% of evaluation papers), and in papers using constructive methods (26\%), artificial evolution (22\%) or declarative programming (19\%). Evaluation is sometimes performed through expressive range analysis (22\% of evaluation papers), sometimes based on experiments with human participants \cite{dimovska2010towards,in2015procedural,antoun2015generating,alvarez2018fostering}, but usually focus on the statistical analysis of computation time (e.g. \cite{etheredge2012fast}), fitness (e.g. \cite{togelius2010towards}), prediction error (e.g. \cite{karavolos2017learning}) and game-specific performance metrics such as map coverage \cite{chowdhury2016exhaustive} or game arcs \cite{de2018evolving}.

\section{Past and Future Trends}\label{sec:trends}
Following the quantitative analysis of different papers' topics, target games, algorithms, etc. the most popular aspects of this corpus will be tracked in terms of their prevalence in different years. Due to the small number of papers published every year, the analysis uses the ratio of relevant papers over all papers published at that year's PCG workshop. The visualization includes a trendline based on linear regression.

\begin{figure}[t]
\centering
\includegraphics[width=0.45\textwidth]{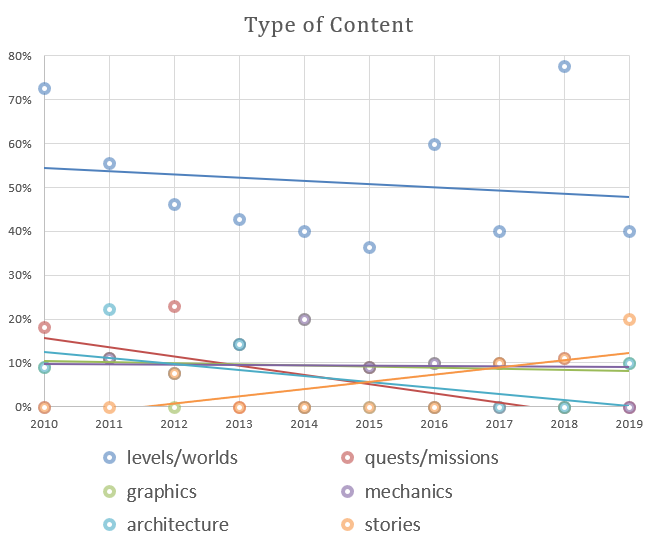}
\caption{The most prominent types of content generated over the years of published PCG papers.}
\label{fig:domains_over_time}
\end{figure}

Fig.~\ref{fig:domains_over_time} shows the most prominent types of content generated, and how they fluctuate across the decade examined. Level generation remains consistently high and fairly steady. Quest generation and architecture had been active topics early on but have been largely absent after the first four years of the workshop, while the generation of mechanics and graphics seems fairly stable in their--somewhat low--prominence. Interestingly, there is an upward trend in narrative generation with recent work at the PCG workshop focusing on aspects of stories \cite{a.2012sharing,grinblat2017subverting,sullivan2018tarot-based,miller2019stories,kreminski2019cozy}; possible reasons for this stronger focus on narrative are discussed in Section \ref{sec:discussion}.

\begin{figure}[t]
\centering
\includegraphics[width=0.45\textwidth]{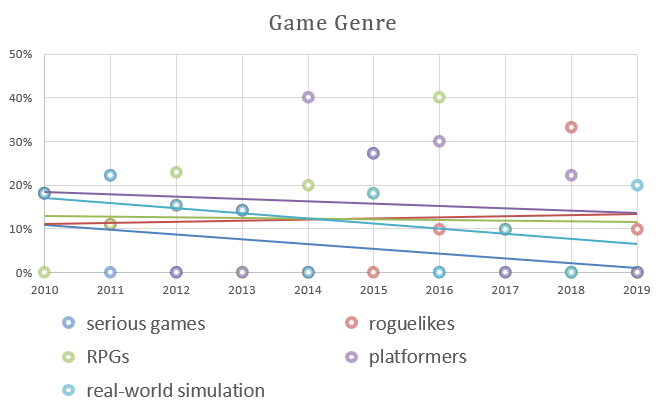}
\caption{The most prominent game genres targeted for generation over the years of published PCG papers.}
\label{fig:genre_over_time}
\end{figure}

Fig.~\ref{fig:genre_over_time} shows the fluctuation in terms of the most prominent game genres targeted. The high fluctuations are evidence that researchers tend to focus on the algorithm or the type of content generated, while the game genre is mostly based on what's available or convenient. Perhaps the only major trend from Fig.~\ref{fig:genre_over_time} is the decrease in research interest on PCG for serious games. There is also a downwards trend for platformer games and an upwards trend for roguelikes, although the small percentage of papers for both seems to indicate that these trends largely depend on available testbeds and personal preferences of contributing researchers. While not in Fig.~\ref{fig:genre_over_time}, worth noting is a recent increase in crafting games, spurred by the Generative Design in Minecraft competition introduced in 2018 \cite{salge2018generative} which provides a challenge and a codebase for interested participants. Crafting games are the focus of 11\% of PCG papers in the last two years, with no presence in prior years.

\begin{figure}[t]
\centering
\includegraphics[width=0.45\textwidth]{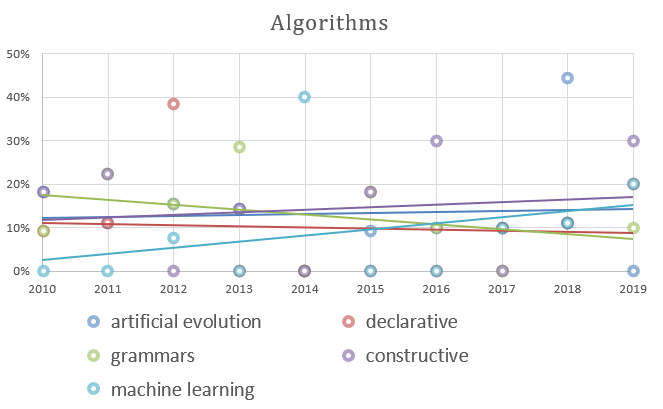}
\caption{The most prominent families of algorithms used in published PCG papers over the years.}
\label{fig:algorithms_over_time}
\end{figure}
Fig.~\ref{fig:algorithms_over_time} shows the major families of algorithms applied for generation in the PCG workshop papers of the last decade. As with game genre, the low overall ratio for each algorithm leads us to believe that the choice of algorithm was often due to the personal preference (or existing codebase) of each researcher; however, some trends can be detected. Despite---or because of---their simplicity, constructive methods seem to be used steadily with no sign that they are falling out of favor. A previous surge in academic interest for declarative approaches seems to be dwindling (with exceptions), and grammars face a similar downward trend from its early popularity in 2010-2013. Artificial evolution seems to be a steady, popular algorithmic approach with a slight upwards trend. As noted in Section \ref{sec:algorithms}, machine learning (ML) is the most upwards trending algorithmic method with 4 of 7 ML papers published from 2017 onwards. More importantly, all four of these latest ML papers combine machine learning with artificial evolution \cite{karavolos2018pairing,hyunjin2019design-centric} or declarative programming \cite{karth2017wavefunctioncollapse,karth2019addressing}. The heightened interest of PCG workshop publications in machine learning are likely attributed to recent successes in deep learning more broadly, the availability of easy-to-use codebases for machine learning, and a game-specific academic interest in PCG via Machine Learning \cite{summerville2018pcgml}. Therefore, perhaps the most significant trend of Fig.~\ref{fig:algorithms_over_time} is the increased use of machine learning on its own or in conjunction with PCG mainstays such as artificial evolution or declarative programming.

\begin{figure}[t]
\centering
\includegraphics[width=0.45\textwidth]{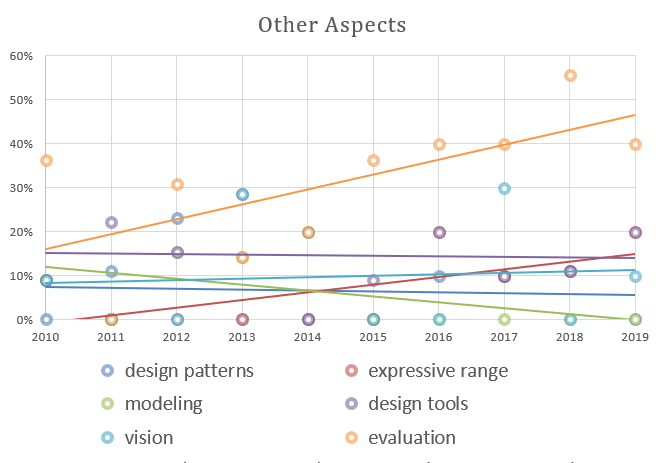}
\caption{Other trends found over the years of published PCG papers.}
\label{fig:other_over_time}
\end{figure}

Finally, Fig.~\ref{fig:other_over_time} tracks how some important aspects of PCG papers highlighted in Section \ref{sec:other} have appeared at different times. It would seem that vision papers remain constant across the years, and workshop organizers should expect at least one vision paper published every year. Design tools and the use of design patterns are similarly retaining their popularity without showing any strong trends upwards or downwards. Interestingly, modeling players and generating personalized content was trending upwards in the first few years but has lost its appeal in the last four years of the workshop. Finally, it is promising that a steadily increasing amount of PCG papers contain a rigorous quantitative evaluation (i.e. not only a qualitative discussion on a few sample outcomes). While in the last two years of PCG papers evaluation is still absent in the majority of papers, the upward trend is irrefutable. Perhaps due to the increase in papers that perform evaluation, more and more papers use the notion of expressive range for the purposes of visualizing the generator's output.

\section{Discussion}\label{sec:discussion}

This paper attempted to categorize the 95 papers published at the PCG workshop in its first decade, as well as study the historical trends in topics covered within these papers. In many cases, classification was difficult and perhaps error-prone; as noted in Section \ref{sec:generated_content} it is difficult to label e.g. a \emph{Minecraft} (Mojang, 2011) settlement or building as level generation or parametric architecture. Boundaries between game genres are fuzzy for commercial games, and even fuzzier when prototype and simplified games are being generated. Similar issues arise when classifying algorithms, as semantic-based generation is sometimes following a declarative approach \cite{lopes2013mobile} and at other times a constructive one \cite{teng2017a}. Section \ref{sec:algorithms} highlighted the dangers of labeling everything as constructive if it does not test the generated artefact \cite{togelius2011searchbased}, and we opted for a stricter definition for constructive algorithms. Due to these issues of ambiguity, many papers ended up without a classification in terms of algorithms even when they included technical details of their generators. Although some readers (or papers' authors) may disagree with certain labels, the general trends found in the corpus of 95 papers are fairly indicative of the trends in PCG research.

Another caveat in this analysis is the nature of the workshop itself. Focusing on the 95 workshop papers rather than e.g. the complete PCG literature published in conferences such as FDG (of which the PCG workshop is part), journal papers or books, is bound to draw conclusions topical of the workshop microcosm. Workshop papers typically focus on algorithms that are work-in-progress or at a prototype stage, or invite a discussion on philosophical issues and new ideas. The large number of vision papers at the PCG workshop should thus not be surprising, and similarly the relative absence of experimental validation. In contrast, journal papers on topics related to PCG---e.g. in the IEEE Transactions on Games---include a robust quantitative evaluation. Finally, the fact that the workshop took place in different locations every year (see Table \ref{tab:overview}) combined with the small community of researchers means that some of the trends can be attributed to lack of participation for a year or a shift in researchers' interests. PCG workshop authors follow their own trajectories (e.g. Joris Dormans organized 3 workshops and published 4 papers from 2010 to 2013 but has since moved to industry), and some iterations of the workshop had larger participation from some research groups than others. Keeping these caveats in mind, we argue that the 10 years of the PCG workshop are indicative of the broader trends in the PCG research field. Our reasoning is that the PCG workshop has been running for longer than other PCG-related academic venues, and its community has inspired most of the advancements in PCG of the last decade including the PCG book \cite{shaker2016procedural}, special issues, PROCJAM, and workshops such as Experimental AI in Games and Computational Creativity and Games. While these larger movements were occurring, the PCG workshop continued to run and act as a node for PCG researchers to convene and co-ordinate. 

On the other hand, we must admit that PCG research itself is a microcosm of the broader movements within AI research, and thus the above patterns may not reflect broader AI trends. A mundane observation is that there are other specialized workshops or conferences where topics targeted by a few PCG workshop papers are nigh-ubiquitous: for instance, the Intelligent Narrative Technologies hosted by the Artificial Intelligence and Interactive Digital Entertainment conference focuses heavily on generating narrative. The International Conference on Computational Intelligence in Music, Sound, Art and Design (EvoMusArt) and the International Conference on Computational Creativity focus heavily on generated graphics and audio, usually beyond games. These publication venues likely follow their own trends in terms of the algorithms that are currently rising in prominence. Beyond the PCG workshop (and games more broadly), the use of deep learning has been applied for a broad range of generative tasks, with an especially strong focus on graphics \cite{Chen2018CartoonGANGA,Gatys2016ImageST} as well as music \cite{Hadjeres2017DeepBachAS} and text \cite{Vaswani2017AttentionIA}. Instead, papers using deep learning published at the PCG workshop exclusively focus on level design tasks. Therefore, PCG research understandably focuses on the game-specific concerns of generation and its trends may differ from broader movements even while it is still affected by them (e.g. the prevalence of deep learning).

Based on the analysis carried out in this paper and a more general view of research in procedural content generation (beyond the workshop itself), it is evident that level generation remains a popular and ``easy'' goal for PCG. This can be attributed to the fact that the game industry initially introduced PCG for level generation in games such as \emph{Rogue} in the 1980s and level generation remains by far the most popular industry application of PCG today. From a practical perspective, moreover, level generation makes sense as games often rely on new levels to increase replayability and variety. Finally, level generation is easier due to the availability of more codebases (produced by researchers, game developers, and enthusiasts alike) and the prevalence of level generation competitions. The generation of complete games, while a popular vision in 2012-2015 \cite{treanor2012game-o-matic:,gustafsson2013data,togelius2014characteristics,a.2015data} has not quite taken off. While complete game generation is being attempted in small clusters \cite{Cook2018RedesigningCC,liapis2019orchestrating}, the drop in interest likely comes from the lack of a big ``success story'' either in academia or the game industry. Another reason is the difficulty that a researcher must face when starting out on game generation in a non-toy domain given the lack of codebases, tutorials or assets. Narrative remains a strong direction for PCG research, likely encouraged by the increased interest on interactive narrative commercially (e.g. with LudoNarraCon) and academically (e.g. with the International Conference on Interactive Digital Storytelling conference and the Intelligent Narrative Technologies workshop).

In terms of algorithms, machine learning seems to be the up-and-coming trend in PCG research. This is hardly surprising given the widespread ``success stories'' and debate \cite{McCormack2019AutonomyAA} on art created via deep learning \cite{Goodfellow2014GenerativeAN,Gatys2016ImageST}. The broader concept of PCG via Machine Learning \cite{summerville2018pcgml} has already driven research in the PCG workshop \cite{karth2019addressing}, while more established PCG algorithms such as artificial evolution or declarative programming are moving towards the integration of machine learning for content initialization \cite{karth2019addressing}, fitness prediction \cite{karavolos2018pairing} and more. It is expected, therefore, that PCG research in deep learning will continue trending upwards, probably combining deep learning with other popular PCG algorithms. 

As noted in Section \ref{sec:target_games}, game genres usually favored by PCG researchers are largely due to personal preference, nostalgia, and convenience. In terms of the latter, the \emph{Super Mario Bros.} level generation competition \cite{shaker2011competition} has largely driven the popularity of platformer games for PCG, not only due to the extensive benchmarks established over the years but also due to the availability of a straightforward generation codebase and efficient gameplaying agents \cite{togelius2010marioai}. The entry level requirements for generation in \emph{Super Mario Bros.} are lower than any other, and researchers can quickly put together something. It is surmised that similar levels of support by both the competition organizers and the community can also drive PCG research to other, overlooked genres. In that regard, \emph{Minecraft} is likely be the next focus point of PCG research, in part due to the GDMC competition \cite{salge2018generative}, the research by Microsoft on \emph{Minecraft} gameplaying agents \cite{hofmann2019minecraft}, and its associated MineRL competition \cite{guss2019the}. There is extensive evidence that academic and practitioner interest hinges on an intuitive codebase as well as an activity that engages the community and produces reusable assets (agents, content, code) that can help future endeavors. Game generation, for which there has been dwindling interest, could thus be reinvigorated with any combination of benchmarks, competitions, easy-to-use codebases, tutorials etc. The same applies to any under-studied game genre, such as stealth games. At the minimum, researchers open-sourcing their generators would help drive future research in their preferred genre. However, building communities through competitions, tutorials, or collaborations with game companies would better act as catalysts to invigorate PCG research on topics that have not yet taken off.

With PCG research moving towards a more established if multi-disciplinary \cite{whitehead2017art} field of study, it is heartening that more attention is given to methodically evaluating and reporting on generators' outcomes. Papers in the PCG workshop have been steadily improving their technical accuracy through quantitative tests with multiple runs and/or human users. Published papers on content generation in journals and conferences have similarly found ways of reporting the quality and diversity of their output. The concept of expressive range \cite{smith2010analyzing} is particularly suited for evaluation of PCG on both dimensions, and a number of enhancements have already been proposed \cite{cook2016danesh:,Summerville2018expressiverange}. There is already an upwards trend in the application of expressive range for PCG evaluation, both in the PCG workshop and beyond. It would be beneficial---at least to the scientific facet of the PCG research field---to establish, contribute to and follow standards of evaluation such as the expressive range of the generator.

\section{Conclusion}\label{sec:conclusion}

This paper celebrates the first decade of the workshop on Procedural Content Generation by providing an analysis of the 95 papers published during these ten years. The paper explored the types of content generated, the game genres, and the algorithms which were favored historically by the PCG workshop community. Drawing from the fluctuations of different topics over the years, the paper identified some trends in the past and the future of the workshop and discussed them in the broader context of PCG research. While the publications at the PCG workshop are a subset of publications on the topic of generation in journals, conferences, books etc., the consistency of the corpus allows us to draw conclusions that are generally in line with the broader academic trends in PCG. The paper is intended primarily as a historical overview, and secondarily as a view towards the future, the challenges and necessary steps for the next ten years of PCG research.

\begin{acks}
The author would like to thank the paper's reviewers for their insightful comments. While not all comments could be addressed in this paper, there is much that needs to be discussed on the future of the PCG workshop and the field at large.
\end{acks}

\bibliographystyle{ACM-Reference-Format}
\bibliography{tenyears,pcg_papers} 

\end{document}